\documentclass{article} 
\usepackage{times}
\usepackage{nips15submit_e}
\usepackage{url}
\usepackage{amsfonts}
\usepackage{amsmath}
\usepackage[colorinlistoftodos,prependcaption]{todonotes}
\usepackage{pgf}
\usepackage{subfig}
\usepackage{booktabs}
\usepackage{resizegather}
\usepackage{setspace}

\usepackage{natbib}

\newcommand\blfootnote[1]{%
  \begingroup
  \renewcommand\thefootnote{}\thanks{#1}%
  \addtocounter{footnote}{-1}%
  \endgroup
}

\title{Manifold Alignment Determination:\\ finding correspondences across different data views}

\author{Andreas Damianou
\\
  Amazon.com\\
  {\tt damianou@amazon.com}
  \And {Neil D. Lawrence}\\
  Amazon.com \& University of Sheffield\blfootnote{Work done while A. Damianou and N. Lawrence were at the University of Sheffield.}\\
  {\tt lawrennd@amazon.com}
  \And {Carl Henrik Ek}\\
  University of Bristol, Bristol, UK\\
  {\tt carlhenrik.ek@bristol.ac.uk}
}

\newcommand{\bfY}{\mathbf{Y}}
\newcommand{\bfy}{\mathbf{y}}
\newcommand{\bfX}{\mathbf{X}}
\newcommand{\bfx}{\mathbf{x}}

\nipsfinalcopy 

\begin{document}

\maketitle

\begin{abstract}
  We present Manifold Alignment Determination (MAD), an algorithm for learning alignments between data points from multiple views or modalities. The approach is capable of learning correspondences between views as well as correspondences between individual data-points. The proposed method requires only a few aligned examples from which it is capable to recover a global alignment through a probabilistic model. The strong, yet flexible regularization provided by the generative model is sufficient to align the views. We provide experiments on both synthetic and real data to highlight the benefit of the proposed approach.
\end{abstract}

\section{Introduction}
Multiview learning is a strand of machine learning which aims at consolidating multiple corresponding views in a single model. The underlying idea is to exploit correspondence between the views to learn a shared representation. A very common scenario is when the views are coming from different modalities. As an example, assume that we are observing a conversation and separately recording the audio and the video of the interaction. Audio and video are two disparate modalities, meaning that it is non-trivial to compare the audio signal to the video and vice-versa. For example, it is nonsensical to compute the Euclidean distance between a video and an audio frame. However, by exploiting the fact that the signals are aligned temporally, we can obtain a correspondence that can be exploited to learn a joint representation for both signals. This shared, multiview model now provides a means to compare the views, allowing us to transfer information between the sound and video specific modalities and enabling joint regularization when performing inference.

The scenario explained above is based on the assumption that the data-points in each view are aligned, i.e. that for each video frame there is a single corresponding sound snippet. This is a strong assumption, as different alignments would lead to completely different representations. In many scenarios we do not know the correspondence between the data-points and this uncertainty should be included in the multiview model. Furthermore, discovering an efficient explicit alignment is in many scenarios the ultimate task, for example when trying to solve correspondence problems. However, discovering an alignment means that we need to search over the space of all possible permutations of the data-points. In practice this is infeasible, as the number of possible alignments grows super-exponentially with the number of data-points.

In this paper we present an algorithmic solution that circumvents the problem of having to rely on only pre-aligned data for the purpose of multiview learning. Rather, we exploit the regularization of a probabilistic factorized multiview latent variable model which allows us to formulate the search as a bipartite-matching problem. We propose two different approaches: one myopic, which aligns data-points in a sequential (iterative) fashion and a nonmyopic algorithm which produces the optimal alignment for a batch of data-points. In our current implementation, both methods rely on a small number of initial aligned instances to act as a ``prior'' of what an alignment means in the particular scenario. We denote this number by $N_{\text{init}}$. We further denote by $N_*$ the number of data-points to be aligned, and $N = N_{\text{init}} + N_*$, $N_{\text{init}} \ll N_*$. The myopic method developed here scales with $O(N_*)$ while the nonmyopic one scales with $O(N_*^3)$ (but in practice it can be run in less than a minute for a few hundreds of data-points). 

\section{Methodology}
We will now proceed to explain the concept of multiview modelling, describe the probabilistic model and how it can be used for automatic alignment within our algorithm. Observing two views $\mathbf{Y}^{(1)}=\{\mathbf{y}^{(1)}_i\}_{i=1}^N$ and $\mathbf{Y}^{(2)}=\{\mathbf{y}^{(2)}_i\}_{i=1}^N$, where $\mathbf{y}^{(1)}_i\in\mathbb{R}^{D_1}$ and $\mathbf{y}^{(2)}_i\in\mathbb{R}^{D_2}$, we consider both views to have been generated from a shared latent variable $\mathbf{X}=\{\mathbf{x}_i\}_{i=1}^N$ where $\mathbf{x}_i\in\mathbb{R}^Q$. The two views are \emph{implicitly} aligned by a permutation ${\boldsymbol \pi} = \{ \pi_i \}_{i=1}^N$ as follows: 
%
if $\bfy^{(1)}_i$ is generated from $\bfx_i$ and $\bfy^{(2)}_i$ is generated from $\bfX \pi_i$, then $\bfy^{(1)}_i$ and $\bfy^{(2)}_i$ are in correspondence.
Since the generating space and the permutation are unobserved, we wish to integrate them out. That is, we wish to be able to compute the marginal likelihood:
\begin{align}
  p(\mathbf{Y}^{(1)},\mathbf{Y}^{(2)}) = \int_{{\boldsymbol \pi}} p(\boldsymbol \pi) \prod_{i=1}^{N} \int_{\mathbf{x}_i} p(\mathbf{y}^{(1)}_i|\mathbf{x}_i) p(\mathbf{y}^{(2)}_i|\mathbf{X}\pi_i) p(\bfx_i), \label{eq:margLikelihoodPerm}
\end{align}
assuming that the observations are conditionally independent given the latent space. Recently, \citeauthor{Klami:2013bw} \citep{Klami:2013bw} formulated a variational approximation of the above model. The key to solve the challenging marginalization over the permutations comes from combining a uniform prior and an interesting experimental observation of permutations. In \citep{Leskovec:2010wb} the authors sample permutations and show that only a very small number of permutations are associated with a non-zero probability. In \citep{Klami:2013bw} this behaviour is exploited by replacing the integral with a sum over a small subset of permutations that are highly probable.

Here we take a different approach. Rather than assuming completely unstructured observations and a uniform prior over alignments, we assume that we are given a small set of aligned pairs. These pairs are used as ``anchors'' from which we search for the best alignment of the remaining points. To achieve this we do not include the permutation as a part of the model but rather propose an algorithm able to search for permutations given a model already estimated in this small set of given aligned pairs.

\section{Manifold Alignment Determination}
We adopt the Manifold Relevance Determination (MRD) \citep{Damianou:2012wu,damianou2016IBFA} as a generative model. The MRD is a non-parametric Bayesian model for multiview modelling.  It is capable of learning compact representations and its Bayesian formulation allows for principled propagation of uncertainty through the model. This regularization enables learning from small data-sets with many views. The learned latent space consolidates the views with a factorized representation which encodes variations which are shared between the views separately to those that are private. This structure is known as Inter Battery Factor Analysis (IBFA) \citep{Tucker:1958ul}. The IBFA structure means that if two views contain pairs of observations that are completely different it will effectively learn two separate models. In the other end of the spectrum, the two views could be presented in pairs which are in perfect correspondence and contain exactly the same information, corrupted by different kinds of (possibly high dimensional) noise. In the later case, IBFA will learn a completely shared model. Importantly, anything between these two extremes is possible and the model does naturally prefer solutions which are as ``shared'' as possible, as this increases compactness which is in accordance with the model's objective function. This is the key insight motivating our work, as a good alignment is such that the views are maximally shared, i.e. we argue and show that the MRD naturally seeks to align the data. The IBFA structure is also adopted by ``Bayesian Canonical Correlation Analysis'' (BCCA) \citep{Klami:2012tf} which was used as a generative model in \citep{Klami:2013bw}. The MRD and the BCCA model are structurally related, but while BCCA is limited to linear mappings, MRD is formulated in a non-parametric fashion, allowing for very expressive non-linear mappings.

We will now describe the algorithm and the model used to learn alignments in more detail. Inspired by the underlying generative model we  refer to this approach as Manifold Alignment Determination or simply as MAD. Given are two views $\mathbf{Y}^{(1)}$ and $\mathbf{Y}^{(2)}$ as defined above, split by two index sets $A=\{i|i\in\mathbb{Z},i\leq N, \mathbf{y}^{(1)}_i\leftrightarrow\mathbf{y}^{(2)}_i\}$ and $B=\{i|i\in\mathbb{Z},i\leq N,i\notin A \}$ so that $|A|=N_{\text{init}}, |B|=N_*$ and where $\mathbf{y}^{(1)}_i\leftrightarrow\mathbf{y}^{(2)}_i$ means that the points are in correspondence. The task is now to learn a latent representation that reflects the correspondence in set $A$ and then align the set $B$ accordingly. In the MRD model there is a single latent space $\mathbf{X}$ from which the two observation spaces $\mathbf{Y}^{(1)}$ and $\mathbf{Y}^{(2)}$ are generated. The mappings $f^{(1)}$ and $f^{(2)}$ from the latent space are modelled using Gaussian process priors with Automatic Relevance Determination (ARD) \citep{Neal:1996ex} covariance functions. The ARD covariance function contains a different weight for each dimension of its input. Learning these weights within a Bayesian framework, such as in MRD and MAD, allows the model to automatically ``switch off'' dimensions of the latent space for each view independently, thereby factorizing the latent space into shared and private subspaces. In the MAD scenario the model is,
\begin{align}
\label{eq:margLikelihood1}
  p(\mathbf{Y}^{(1)}_A, & \mathbf{Y}^{(2)}_A|\mathbf{w}^{(1)},\mathbf{w}^{(2)}) = \nonumber \\
  & \int_{f^{(1)},f^{(2)},\mathbf{X}}p(\mathbf{F}_A^{(1)}|\mathbf{X},\mathbf{w}^{(1)})p(\mathbf{F}_A^{(2)}|\mathbf{X},\mathbf{w}^{(2)})p(\mathbf{X}) \prod_{\forall i \in A} p(\mathbf{y}_i^{(1)}|\mathbf{f}^{(1)}_i) p(\mathbf{y}_i^{(2)}|\mathbf{f}^{(2)}_i),
\end{align}
where $\mathbf{F}_A = \{ \mathbf{f}_i \}_{\forall i \in A}$ and $\mathbf{w}^{(1)}$ and $\mathbf{w}^{(2)}$ are the aforementioned parameters of the ARD covariance function. Specifically, $w^{(j)}_i$ determines the importance of dimension $i$ when generating view $j$. This is what allows the model to learn a factorized latent space, because as a shared dimension is deemed one that has a non-zero weight for each view and a private dimension is one for which only one view has a non-zero weight. Marginalising the latent space from the model is intractable for general covariance functions, as the distribution over $\mathbf{X}$ is propagated through a non-linear mapping. In \citep{Damianou:2012wu,damianou2016IBFA} a variational compression scheme is formulated providing a lower bound on the log-likelihood, and this approach is also followed in our paper. In other words, we aim to maximize the functional $\mathcal{F}(\mathcal{Q}, \mathbf{w}^{(1)},\mathbf{w}^{(2)}) \le p(\mathbf{Y}^{(1)}_A,\mathbf{Y}^{(2)}_A|\mathbf{w}^{(1)},\mathbf{w}^{(2)})$ with respect to the ARD weights and the introduced variational distribution $\mathcal{Q}$ which governs the approximation. 

Notice that we can equivalently write equation \eqref{eq:margLikelihood1} after the marginalisation of $\mathbf{F}_A$ is performed:
\begin{gather}
\label{eq:margLikelihood2}
  p(\mathbf{Y}^{(1)}_A,\mathbf{Y}^{(2)}_A|\mathbf{w}^{(1)},\mathbf{w}^{(2)}) = \int_{\mathbf{X}}   p(\mathbf{X}) p(\mathbf{Y}^{(1)}_A|\mathbf{X}, \mathbf{w}^{(1)}) p(\mathbf{Y}^{(2)}_A|\mathbf{X}, \mathbf{w}^{(2)}).
\end{gather}
As can be seen, the marginalization of $\mathbf{F}_A$ couples all outputs with index in $A$ within each view. Furthermore, by comparing to equation \eqref{eq:margLikelihoodPerm} we see that in our MAD framework, the explicit permutation of one view is replaced with relevance weights for all views. These weights alone cannot play a role equivalent to a permutation, but the permutation learning is achieved when we combine these weights with the learning algorithms described below.

\subsection{Learning Alignments}
Assume that we have trained our multiview model on a small initial set of $N_{\text{init}}$ aligned pairs $( \mathbf{Y}^{(1)}_A, \mathbf{Y}^{(2)}_A )$. For the remaining \emph{unaligned} $N_*$ data-points $( \mathbf{Y}^{(1)}_B, \mathbf{Y}^{(2)}_B )$ we wish to find a pairing of the data indexed by set $B$ which corresponds to the alignment of the data indexed by $A$. As previously stated, learning is performed by optimizing a variational lower bound on the model marginal likelihood. Since we have an approximation to the marginal likelihood, we can obtain an approximate posterior $\mathcal{Q}$ of the integrated quantities.  In particular, we are interested in the marginal for $\bfX$. We denote this approximate posterior as $q(\mathbf{X})$, and assign it a Gaussian form (the mean and variance of which we learn variationally). Importantly, as each view is modelled by a different Gaussian process, there are two different posteriors for the same latent space:
\begin{equation}
\begin{aligned}
\label{eq:posteriors}
  q_{(1)}(\mathbf{X}_A,\mathbf{X}_B) = \prod_{i \in A} q_{(1)}(\bfx_i) \prod_{i \in B} q_{(1)}(\bfx_i) \approx p(\mathbf{X}_A,\mathbf{X}_B|\mathbf{Y}^{(1)}_A,\mathbf{Y}^{(1)}_B),\\
q_{(2)}(\mathbf{X}_A,\mathbf{X}_B) = \prod_{i \in A} q_{(2)}(\bfx_i) \prod_{i \in B} q_{(2)}(\bfx_i) \approx p(\mathbf{X}_A,\mathbf{X}_B|\mathbf{Y}^{(2)}_A,\mathbf{Y}^{(2)}_B), 
\end{aligned}
\end{equation}
one for each view. The agreement of the two different posteriors provides us with means to align the remaining data. There are several different possibilities to align the data and here we have explored two different approaches which both give very good results. The first approach is myopic and on-line with respect to one of the views, and uses only a limited ``horizon''. This approach is appropriate for the scenario where we observe points sequentially in one of the views, e.g. view $(2)$, and wish to match each incoming point to a set in the other view, e.g. view $(1)$. To solve this task we first compute the posterior for all points in view $(1)$, i.e. $q_{(1)}(\mathbf{X}_A,\mathbf{X}_B)$ using equation \eqref{eq:posteriors}. Then, for each incoming point $\bfy^{(2)}_i, i \in B$, we compute the posterior $q_{(2)}(\mathbf{x}_i,\mathbf{X}_A, \mathbf{X}_{\hat{B}})$, where $\hat{B}$ denotes all the so far obtained points from set $B$. Finally, we find the best matching between the mode of $q_{(2)}(\mathbf{x}_i,\mathbf{X}_A, \mathbf{X}_{\hat{B}})$ and the mode of each marginal $q_{(1)}(\mathbf{x}_i,\mathbf{X}_A), i \in B$. That is, we pair the point with the closest mean of the posterior from the other view. This can be seen a procedure by which we map incoming points $\bfy^{(2)}_i$ to a latent point $\bfx_i$ through the posterior and then performing a nearest neighbour search in the latent space. Importantly, the comparative matching is performed in the \emph{shared} latent space of the two views, thereby ignoring private variations that are irrelevant with respect to the alignment of the views. 

Clearly the above approach is very fast but depends on the order by which we observe the points. Therefore we also explore a nonmyopic approach, where we observe all the data at the same time. This allows us to perform a full bipartite matching by using the Hungarian method \citep{Kuhn:2010cm}. In this scenario, we compute $q_{(1)}(\mathbf{X}_A,\mathbf{X}_B)$ and $q_{(2)}(\mathbf{X}_A,\mathbf{X}_B)$ at the same time, from where we extract the marginals $q_{(1)}(\mathbf{X}_B)$ and $q_{(2)}(\mathbf{X}_B)$. These distributions factorize with respect to data points (equation \eqref{eq:posteriors}), so we can obtain $N_*$ modes for each. These modes are then compared in a $N_* \times N_*$ distance matrix $D_{**}$ (we use the $L_2$ distance). This distance matrix is given as an objective to the Hungarian method. As before, the distance between the modes is only computed in the \emph{shared} dimensions for both views. Notice that under this point of view, MAD is a means of performing a full bipartite match in views which are cleaned from noise and private (irrelevant) variations, are compressed (so the search is more efficient) and, most importantly, are embedded in the same space. Performing the bipartite match in the output space would not be possible, as these spaces are not comparable.

There are many other possibilities to perform the nonmyopic matching. For example, we might know that the data have been permuted by a linear transformation and seek to find the alignment by searching for the Procrustes superimposition. In practice we have used the Hungarian method to match the posteriors and have obtained better results compared to the myopic method, while the myopic approach is perhaps only suitable when we have thousands or millions of data points to match. Therefore, in our experiments we use the nonmyopic approach.

\section{Experiments}

In this section we present experiments in simulated and real-world data in order to demonstrate the intuitions behind our method -- manifold alignment determination (MAD) -- as well as its effectiveness. To start with, we tested the suitability of the MRD backbone as a penalizer of mis-aligned data. We constructed two $20-$dimensional signals $\bfY^{(1)}$ and $\bfY^{(2)}$ such that they contained shared and private information. We then created $K$ versions of $\bfY^{(2)}$, each denoted by $\hat{\bfY}^{(2)}_k$. Here, $k$ is proportional to the degree of mis-alignment between $\bfY^{(2)}$ and $\hat{\bfY}^{(2)}_k$, which is measured by the \emph{Kendall$-\tau$ distance}. Figure 1 illustrates the results of this experiment. Clearly, the MRD is very sensitive in distinguishing the degree of mis-alignment, making it an ideal backbone for our algorithm.

Next, we evaluate MAD in synthetic data. We constructed an input space $\bfX = \left[ \bfX^{(1)} \bfX^{(2)} \bfX^{(1,2)} \right]$, where $\bfX^{(1)}, \bfX^{(2)} \in \mathbb{R}, \bfX^{(1,2)} \in \mathbb{R}^2$. Notice that $\bfX$ only contains orthogonal columns (dimensions), created as sinusoids with different frequency. Two output signals $\bfY^{(1)}$ and $\bfY^{(2)}$ were then generated by randomly mapping from parts of $\bfX$ to $20$ dimensions with the addition of Gaussian noise (in both $\bfX$ and the mapping) with standard deviation 0.1. We experimented with mappings that create only shared information in the two outputs (by mapping only from $\bfX^{(1,2)}$) and with mappings that created private and shared information (by mapping from $\left[ \bfX^{(1,2)} \bfX^{(1)} \right]$ for $\bfY^{(1)}$ and from $\left[ \bfX^{(1,2)} \bfX^{(2)} \right]$ for $\bfY^{(2)}$). We also experimented with linear random mappings and non-linear ones. The later were implemented as a random draw from a Gaussian process and were handled by non-linear kernels inside our model. Figure (1)(b,c,d) shows the very good results obtained by our method and a quantification is also given in Table 1. The size of the initial set used as a guide to alignments is denoted by $N_{\text{init}}$.

\begin{figure}[t]
\label{fig:kendal1}
  \centering
  \subfloat[MRD marginal log-likelihood with increasing mis-alignments]{\includegraphics[width=0.5\textwidth]{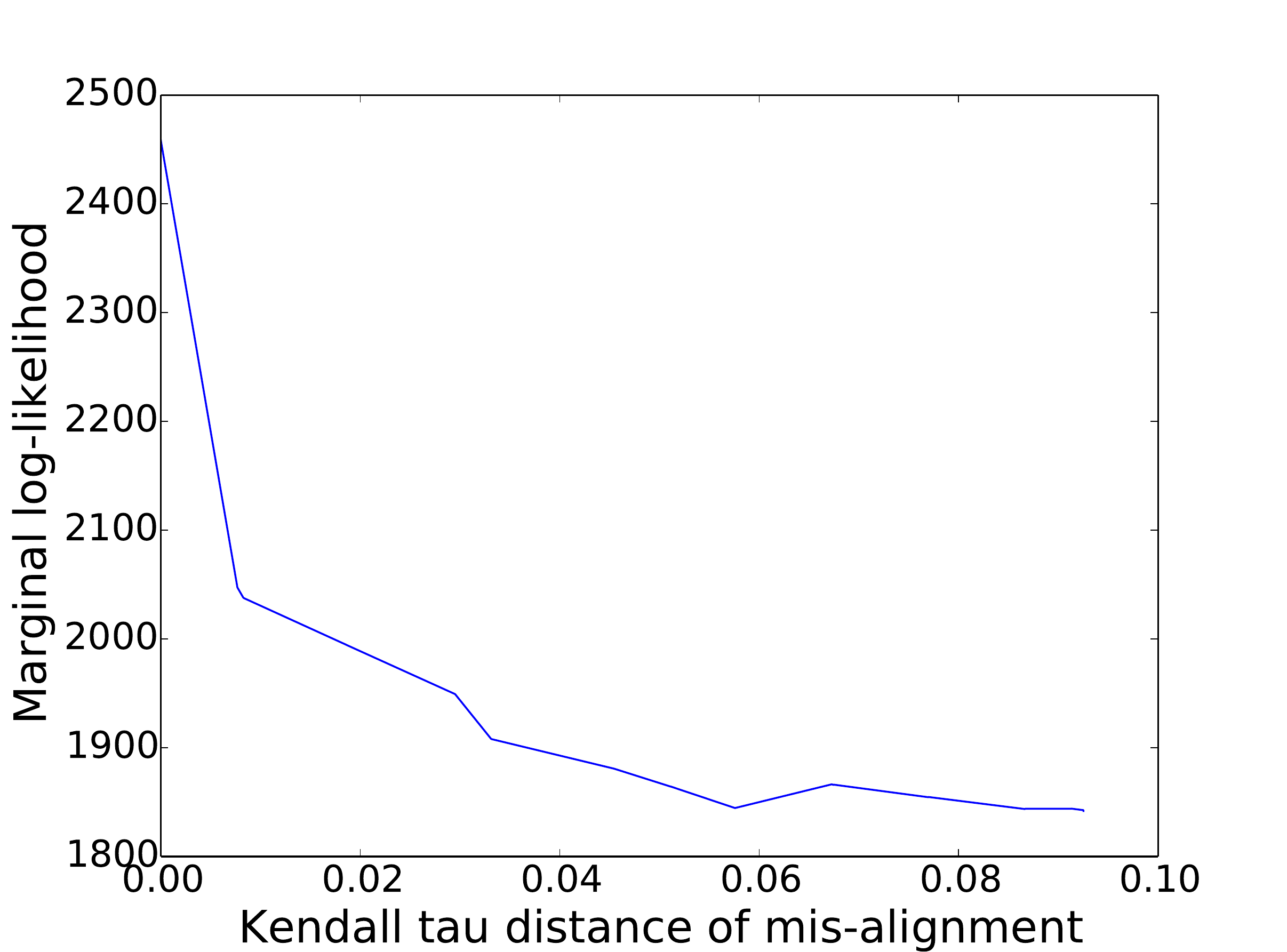}}\label{subfig:kendall} \\
   \subfloat[Linear mapping, fully shared generating space]{\includegraphics[width=0.26\textwidth]{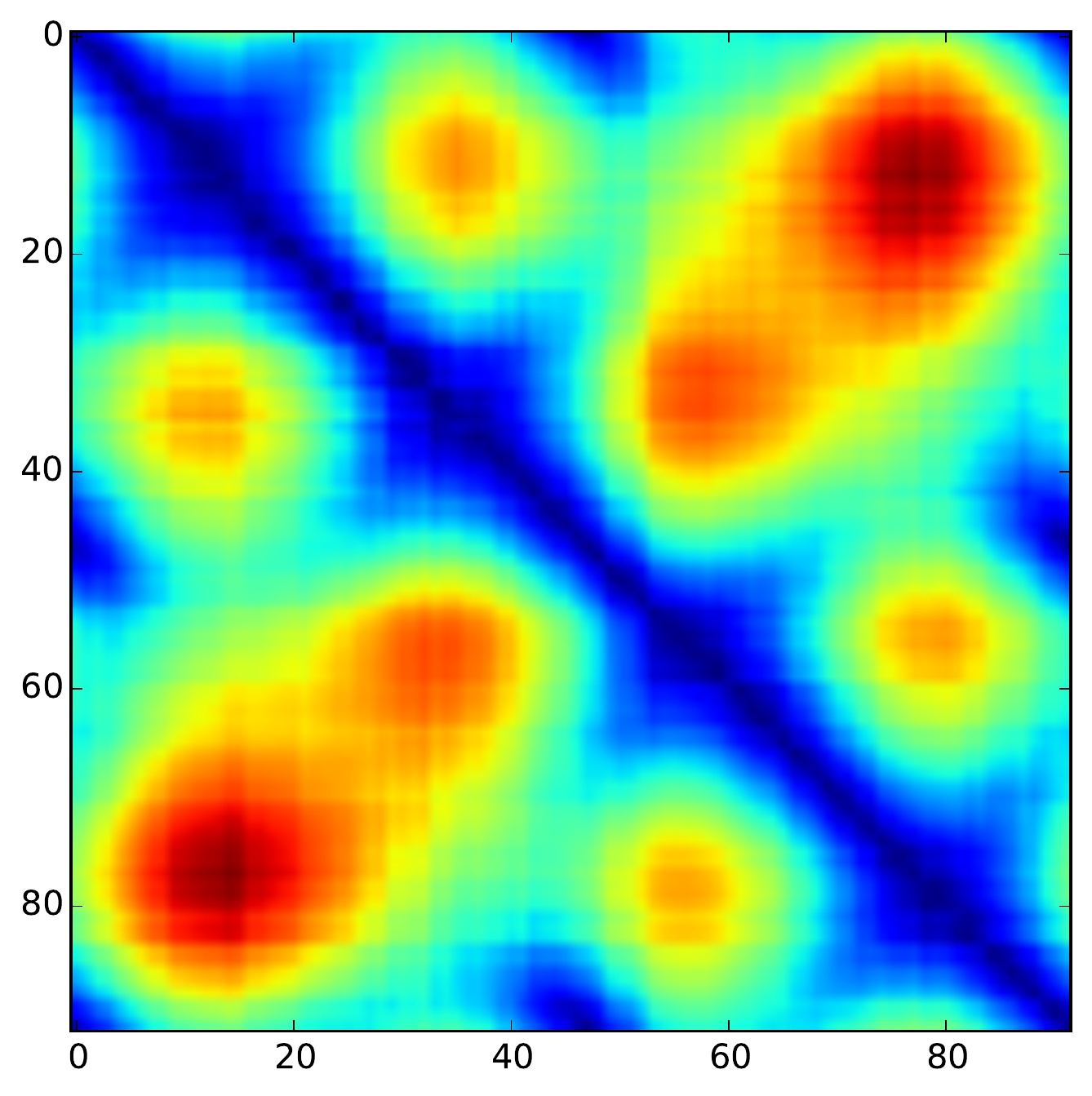}}\label{subfig:toy_linshared}
   \hfill
   \subfloat[Linear mapping, shared/private generating space]{\includegraphics[width=0.26\textwidth]{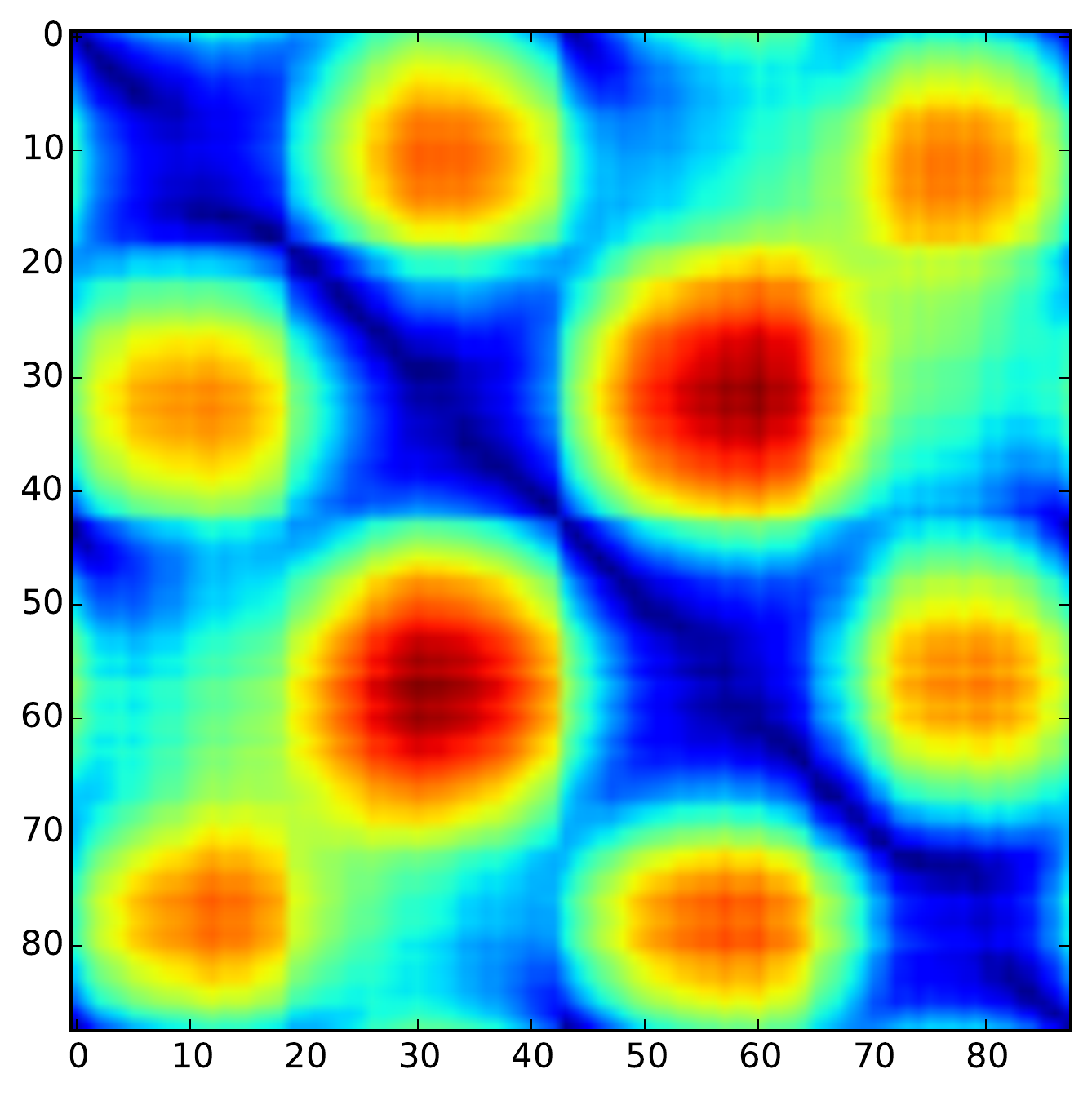}}\label{subfig:toy_linpriv}
   \hfill
   \subfloat[Nonlinear mapping, shared/private generating space]{\includegraphics[width=0.26\textwidth]{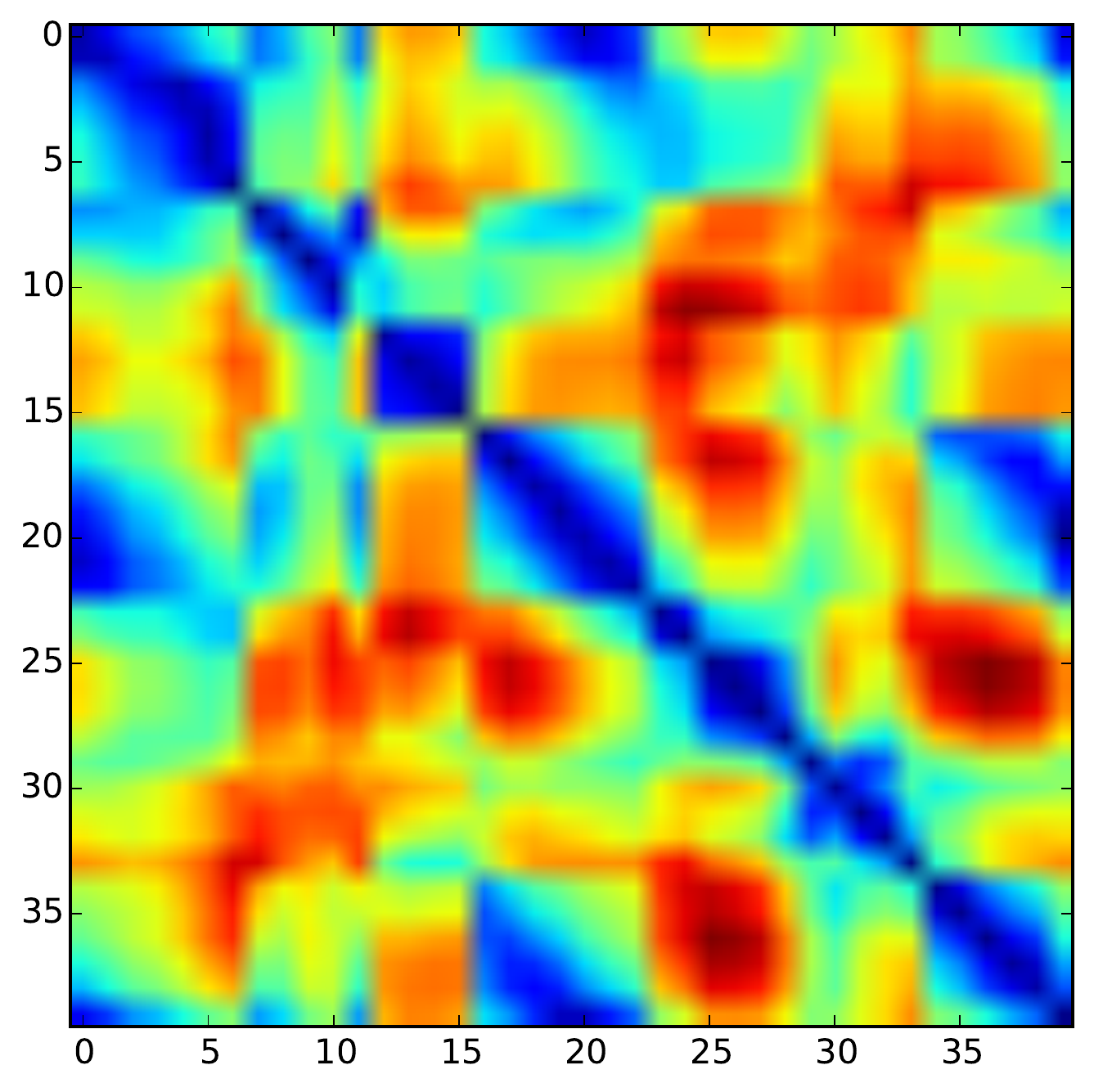}}\label{subfig:toy_nonlinpriv}
   \caption{Results for the toy experiments. Plot (a) demonstrates the suitability of the MRD objective as a backbone of MAD. Plots (b,c,d) visualise the matrix of distances $D_{**}$   of the inferred latent features obtained from $\bfY^{(1)}$ and $\bfY^{(2)}$. The small values around the diagonal suggest that the optimal alignment found is the correct one, i.e. $\bfy^{(1)}_i$ is always matched to $\bfy^{(2)}_j$, where $i-j = \epsilon, \epsilon \rightarrow 0$. Each sub-caption in these figures denotes the way in which the corresponding views were generated for this toy example.}
\end{figure}

We now proceed to illustrate our approach in real data. Firstly, we considered a dataset $\bfY^{(1,2)}$ which contained 4 different human walk motions represented as a set of $59$ coordinates for the skeleton joints (subject 35, motions 1--4 from the CMU motion capture database). We created $\bfY^{(1)}$ and $\bfY^{(2)}$ by splitting in half the dimensions of $\bfY^{(1,2)}$. In another experiment, we selected $100$ handwritten digits representing the number $0$ from the USPS database and similarly created the two views by splitting each image in half horizontally, so that each view contained $16 \times 8$ pixels. The distance matrix $D_{**}$ computed in the inferred latent space, as well as quantification of the alignment mismatch and dataset information can be found in Figure 2 and Table \ref{table:results}. The motion capture data is periodic in nature (due to the multiple walking strands), a pattern that successfully is discovered in $D_{**}$, as can be seen in Figure 2. For the digits demo we also present a visualisation of the alignment, by matching each top-half image from $\bfY^{(1)}$ with its inferred bottom-half from $\bfY^{(2)}$. As can be seen, most of the digits are well aligned along the pen strokes. In a few cases (e.g. the last three digits), the matching between the black and white strokes seems to be inverted, which means that although the alignment is not perfect, it is still following a learned continuity constraint. In these cases, this learned constraint is suboptimal but, nevertheless, satisfactory for the majority of the examples.

\newcommand{\mytab}{
  \begin{tabular}{l|c|c|c|c}
    \toprule
    Data &  $N$ & $N_{\text{init}}$ & Kendall-$\tau$ \\
    \midrule
    Toy, linear, fully shared    & 100 &  8  & 0.001 \\
    \hline
    Toy, linear, shared/private  & 100 &  12 & 0.003  \\
    \hline
    Toy, nonlin., shared/private & 160 &  48 & $2\times 10^{-6}$\\
    \hline
    Motion Capture               & 350 &  175 & 0.21 \\
    \hline
    Digits                       & 100 &  50 & 0.28  \\
    \bottomrule
  \end{tabular}
}

\begin{figure}[t]
\label{fig:kendal_experiments}
  \centering
  \subfloat[]{\includegraphics[width=0.3\textwidth]{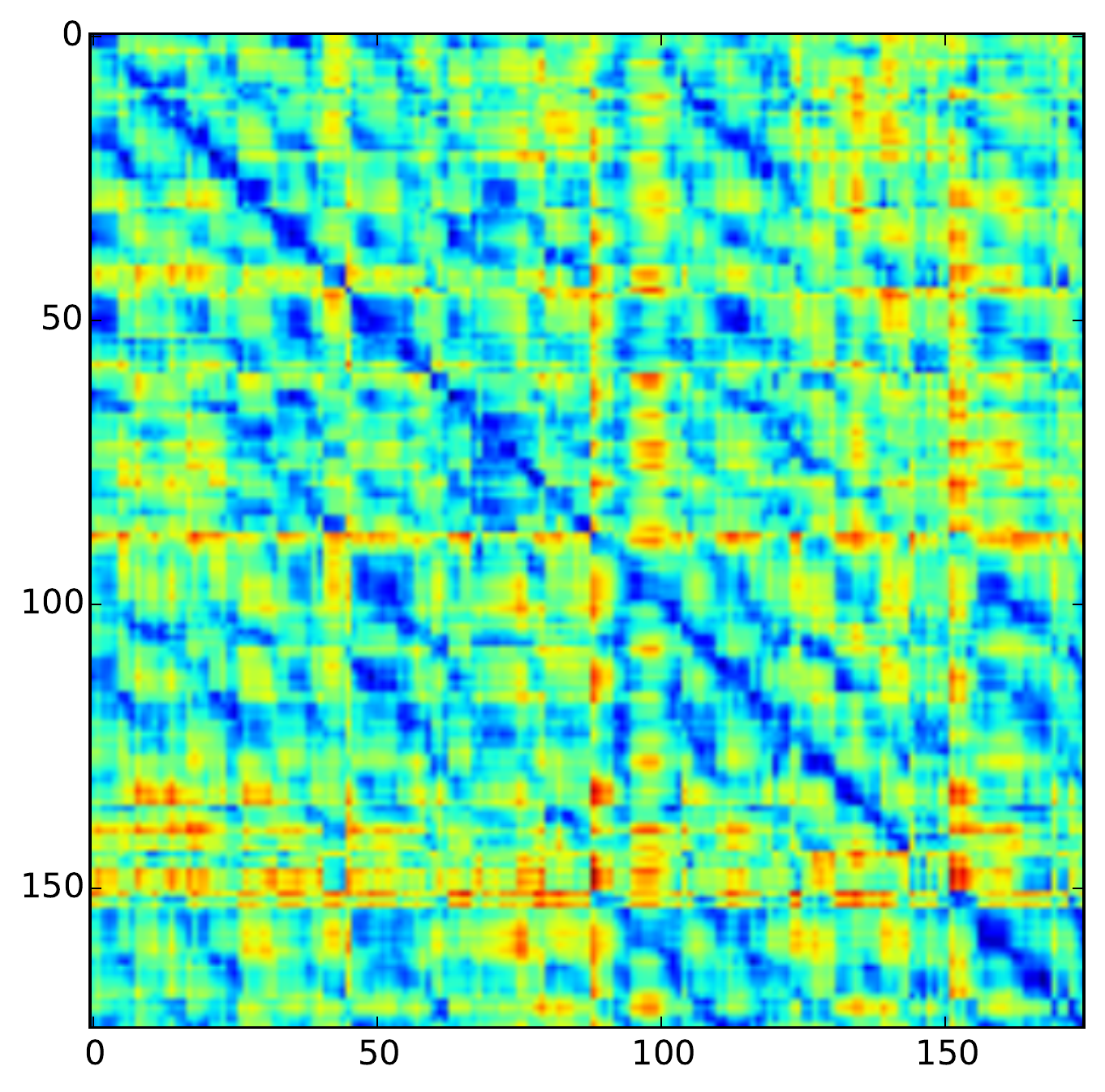}}
  \hspace{20pt}
  \subfloat[]{\includegraphics[width=0.3\textwidth]{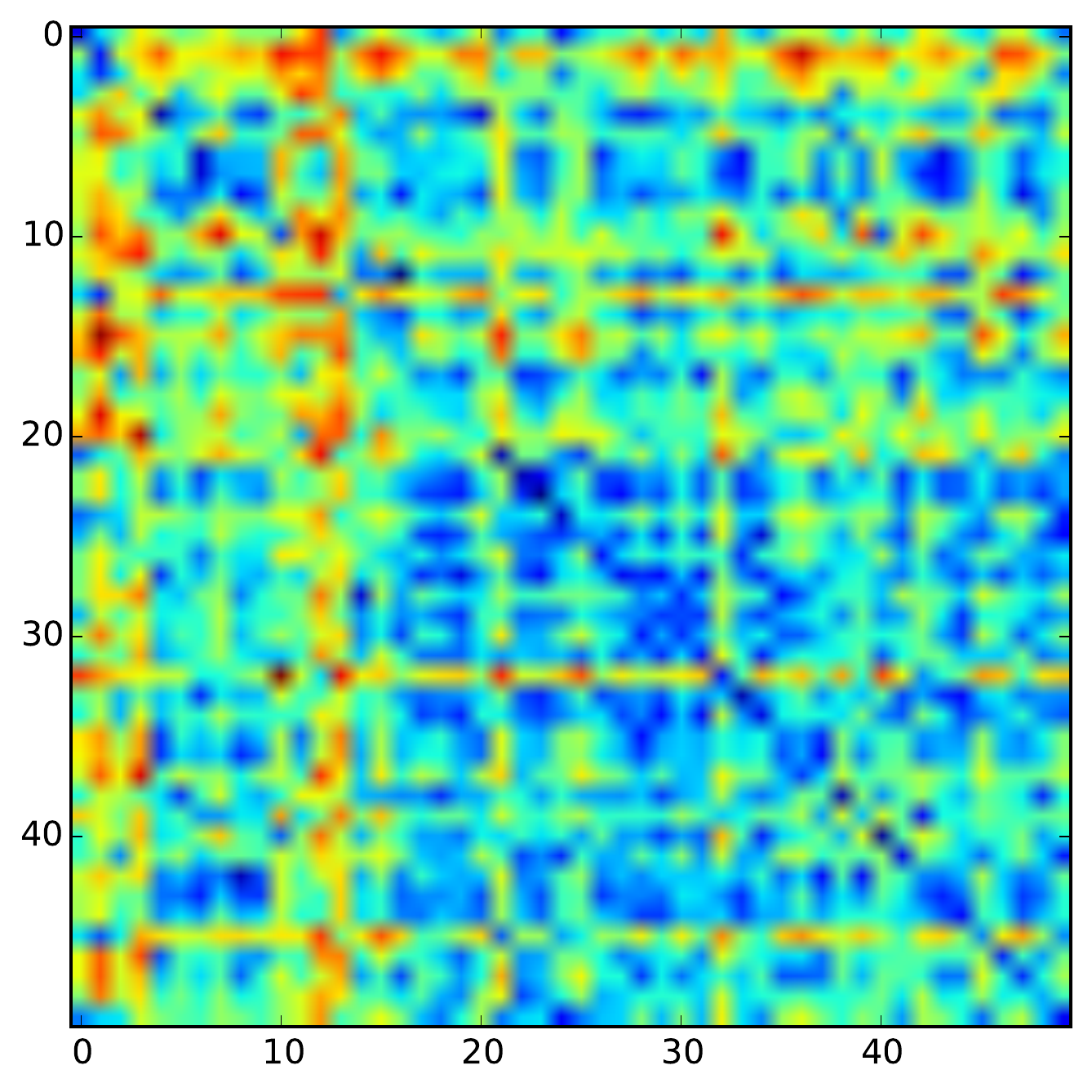}}
  \vspace{20pt}
\caption{The inferred distance matrices $D_{**}$ used to perform the alignments in the input space for the motion capture experiment (left) and the digits experiment (right).}
\end{figure}

\begin{table}[h]
\centering
  \mytab
\caption{Kendall-$\tau$ distance of the data aligned by our algorithm. A random matching would achieve a distance of $\approx 0.5$. Much larger $N$ can be easily handled, but due to time constraints we did not perform such experiments (and also for smaller $N_{\text{init}}$).}
\label{table:results}
\end{table}



\begin{figure}
  \centering
  \includegraphics[width=1\textwidth]{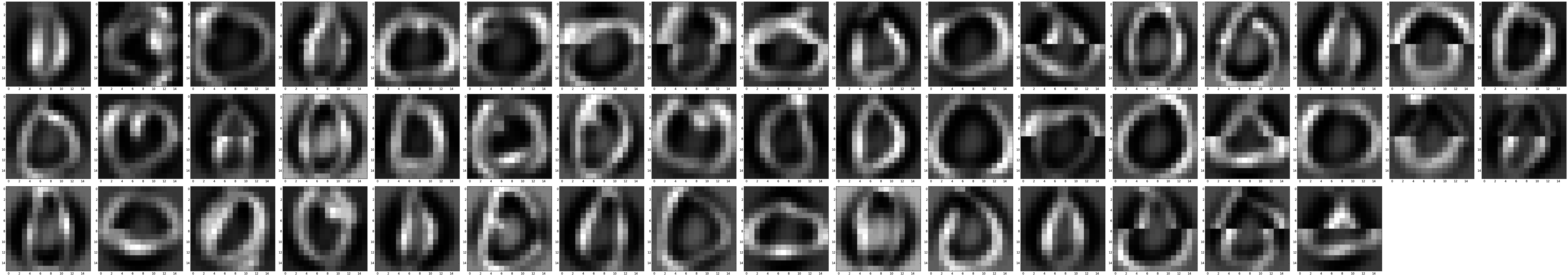}
  \caption{{Matching the top-half with the (aligned by our method) bottom-half parts of the digits.
    }}
\end{figure}

\section{Conclusion}
We have presented an approach for automatic alignment of multiview data. Our key contribution is to show that the combination of a factorized latent space model and Bayesian learning naturally reflects alignments in multiview data. Our algorithm learns correspondences from a small number of examples allowing additional data to be aligned. In future work we will extend the approach to make use of the uncertainty in the approximate latent posterior rather than only using the first mode as in this paper. The full distribution provides means for adding only pairs which the model are relatively certain about and update the model accordingly. Learning alignments is an important (and very challenging) task in multiview learning that we believe needs to be highlighted in order to stimulate further research.

\textbf{Acknowledgements.} While pursuing this work, A. Damianou was funded by the European research project EU FP7-ICT (Project
Ref 612139 ``WYSIWYD'').


\bibliographystyle{plainnat}
\bibliography{2015_nips_ws-blx,carl}

\begin{thebibliography}{8}
\providecommand{\natexlab}[1]{#1}
\providecommand{\url}[1]{\texttt{#1}}
\expandafter\ifx\csname urlstyle\endcsname\relax
  \providecommand{\doi}[1]{doi: #1}\else
  \providecommand{\doi}{doi: \begingroup \urlstyle{rm}\Url}\fi

\bibitem[Damianou et~al.(2016)Damianou, Lawrence, and Ek]{damianou2016IBFA}
Andreas Damianou, Neil~D Lawrence, and Carl~Henrik Ek.
\newblock Multi-view learning as a nonparametric nonlinear inter-battery factor
  analysis.
\newblock \emph{arXiv preprint arXiv:1604.04939}, 2016.

\bibitem[Damianou et~al.(2012)Damianou, Ek, Titsias, and
  Lawrence]{Damianou:2012wu}
Andreas~C. Damianou, Carl~Henrik Ek, Michalis Titsias, and Neil~D. Lawrence.
\newblock {Manifold Relevance Determination}.
\newblock In \emph{International Conference on Machine Learning}, pages
  145--152, June 2012.

\bibitem[Klami(2013)]{Klami:2013bw}
Arto Klami.
\newblock {Bayesian object matching}.
\newblock \emph{Machine learning}, 92\penalty0 (2-3):\penalty0 225--250, 2013.

\bibitem[Klami et~al.(2013)Klami, Virtanen, and Kaski]{Klami:2012tf}
Arto Klami, Seppo Virtanen, and Samuel Kaski.
\newblock {Bayesian Canonical Correlation Analysis}.
\newblock 14:\penalty0 965--1003, April 2013.

\bibitem[Kuhn(2010)]{Kuhn:2010cm}
Harold~W Kuhn.
\newblock {The Hungarian Method for the Assignment Problem.}
\newblock \emph{50 Years of Integer Programming}, \penalty0 (Chapter
  2):\penalty0 29--47, 2010.

\bibitem[Leskovec et~al.(2010)Leskovec, Chakrabarti, Kleinberg, Faloutsos, and
  Ghahramani]{Leskovec:2010wb}
Jure Leskovec, Deepayan Chakrabarti, Jon Kleinberg, Christos Faloutsos, and
  Zoubin Ghahramani.
\newblock {Kronecker Graphs: An Approach to Modeling Networks}.
\newblock \emph{The Journal of Machine Learning Research}, 11:\penalty0
  985--1042, March 2010.

\bibitem[Neal(1996)]{Neal:1996ex}
Radford~M Neal.
\newblock \emph{{Bayesian Learning for Neural Networks}}, volume~8.
\newblock New York: Springer-Verlag, 1996.

\bibitem[Tucker(1958)]{Tucker:1958ul}
Ledyard~R Tucker.
\newblock {An Inter-Battery Method of Factory Analysis}.
\newblock \emph{Psychometrika}, 23, June 1958.

\end{thebibliography}

\end{document}